\documentclass[conference,a4paper]{IEEEtran}
\IEEEoverridecommandlockouts

\usepackage[hidelinks,hypertexnames=false]{hyperref}
\usepackage[cmex10]{amsmath}
\usepackage{amssymb,amsfonts}
\usepackage{dblfloatfix}

\usepackage[ruled,vlined]{algorithm2e}
\usepackage{graphicx}
\usepackage{float}
\graphicspath{{Figures/PDF/}{Figures/PNG/}}

\usepackage{booktabs}
\usepackage{multirow}
\usepackage{siunitx}
\usepackage{xcolor}
\usepackage[numbers,compress]{natbib}
\usepackage{texnames}
\usepackage{bm,bbm}
\usepackage{orcidlink}
\usepackage{times}

\makeatletter
\let\UMBenchOldThebibliography\thebibliography
\let\endUMBenchOldThebibliography\endthebibliography
\renewenvironment{thebibliography}[1]{%
  \fontsize{10pt}{10.5pt}\selectfont
  \UMBenchOldThebibliography{#1}%
  \setlength{\itemsep}{0.2pt}%
  \setlength{\parsep}{0pt}%
  \setlength{\parskip}{0pt}%
}{\endUMBenchOldThebibliography}
\makeatother

\begin{document}



\title{\uppercase{BENCHMARKING FINE-GRAINED VISUAL REASONING FOR REMOTE SENSING IN URBAN MANAGEMENT}}



\author{	
    \IEEEauthorblockN{Jiawei Zhang\textsuperscript{1}, Rizhao Cai\textsuperscript{2}, Dayan Guan\textsuperscript{1}}
    \IEEEauthorblockA{\textsuperscript{1}School of Electronics and Information Engineering, Harbin Institute of Technology, 150000 Harbin, China.\\
	\textsuperscript{2} Huawei International PTE LTd, 486066 Singapore, Singapore.
    }
    \IEEEcompsocitemizethanks{
    \IEEEcompsocthanksitem Corresponding authors: Dayan Guan.
    }
}

\maketitle


\begin{abstract}
Fine-grained visual reasoning is essential for remote sensing in urban management, yet remains underexplored in the evaluation of multimodal large language models (MLLMs). We introduce UMBench, a fine-grained urban remote sensing benchmark built from unmanned aerial vehicle (UAV) images
for multiple urban management tasks such as traffic monitoring, urban construction, environmental quality, and disaster warning. Evaluations on UMBench reveal that current MLLMs often rely on scene-level priors and overlook fine-grained visual evidence. To address this issue, we further propose a novel evidence-first chain-of-thought method that directs models to inspect fine-grained visual evidence in a training-free manner. Experiments show that our method can improve the performance of both open-source and commercial MLLMs on UMBench.


\end{abstract}

\begin{IEEEkeywords}
	Multimodal Large Language Models, Training-free Method, Remote Sensing, Urban Management.
\end{IEEEkeywords}

\section{INTRODUCTION}



Remote sensing (RS) has become an important information source for urban management, supporting a range of applications, including urban sustainability monitoring, disaster risk management, and observation of diverse urban environments \cite{kadhim2016advances, ghaffarian2018remote, chen2024remote}. In these applications, effective image understanding should not stop at scene-level recognition, but should support task-specific reasoning based on fine-grained visual evidence \cite{jiang2025eaglevision, li2026provg}. Since such reasoning tasks are often task-dependent and difficult to cover with a fixed set of predefined labels, MLLMs offer a flexible interface for interpreting RS images through natural-language queries and linking fine-grained visual evidence to task-specific reasoning. However, fine-grained visual reasoning in urban management remains underexplored in MLLM evaluation.

Existing RS evaluation benchmarks primarily focus on scene-level visual capabilities \cite{li2024vrsbench}. However, models may rely on scene-level priors, inferring answers from overall scene context or common object co-occurrence rather than from fine-grained visual evidence, a failure mode related to object hallucination in vision-language models \cite{li2023evaluating}. For example, scene-level questions may ask whether an image contains a road, while fine-grained questions ask whether a vehicle crosses a lane line or whether local cracks exist. Traditional datasets, such as RSVQA \cite{lobry2020rsvqa}, mainly evaluate scene classification or generic question answering, where many answers can be inferred from scene-level information without explicitly testing fine-grained visual evidence. Even high-resolution benchmarks, such as FloodNet \cite{rahnemoonfar2021floodnet}, primarily evaluate scene classification or generic question answering, focusing on scene-level understanding or binary classification. Existing evaluations rarely test the intersection of fine-grained visual inspection and logical reasoning. In real-world RS applications, targets can be minute, ambiguous, and embedded in complex backgrounds. Current evaluations do not test whether an MLLM discerns these fine-grained visual details or guesses from scene-level priors.

To address this gap, we introduce UMBench, an Urban Management Benchmark for fine-grained visual reasoning in complex RS scenarios. It is built from 583 UAV images and contains 2,733 expert-curated QA pairs spanning traffic monitoring, urban construction, environmental quality, and disaster warning. UMBench tests whether models can combine fine-grained visual inspection with reasoning instead of relying on scene-level priors.

To reduce reliance on scene-level priors, we propose a training-free evidence-first chain-of-thought (CoT) prompting method \cite{wei2022chain}. The method guides models to inspect fine-grained visual evidence before making a final choice. Since it requires no model fine-tuning or access to model weights, it can be applied to both open-source MLLMs and commercial MLLMs. Experiments on representative Qwen-VL, InternVL, and commercial models show that the proposed method improves performance across different access settings.

The main contributions of this work are summarized as follows. 
\begin{itemize}
    \item We propose UMBench, the first urban remote sensing benchmark for fine-grained visual reasoning in urban management.
    \item We design a novel training-free method to guide MLLMs to focus on fine-grained visual evidence rather than scene-level priors.
    \item The designed method is applicable to both open-source and commercial MLLMs due to its training-free property.
\end{itemize}

\section{UMBENCH DATASET}

\begin{figure*}[t]
\centering
\includegraphics[width=0.96\textwidth]{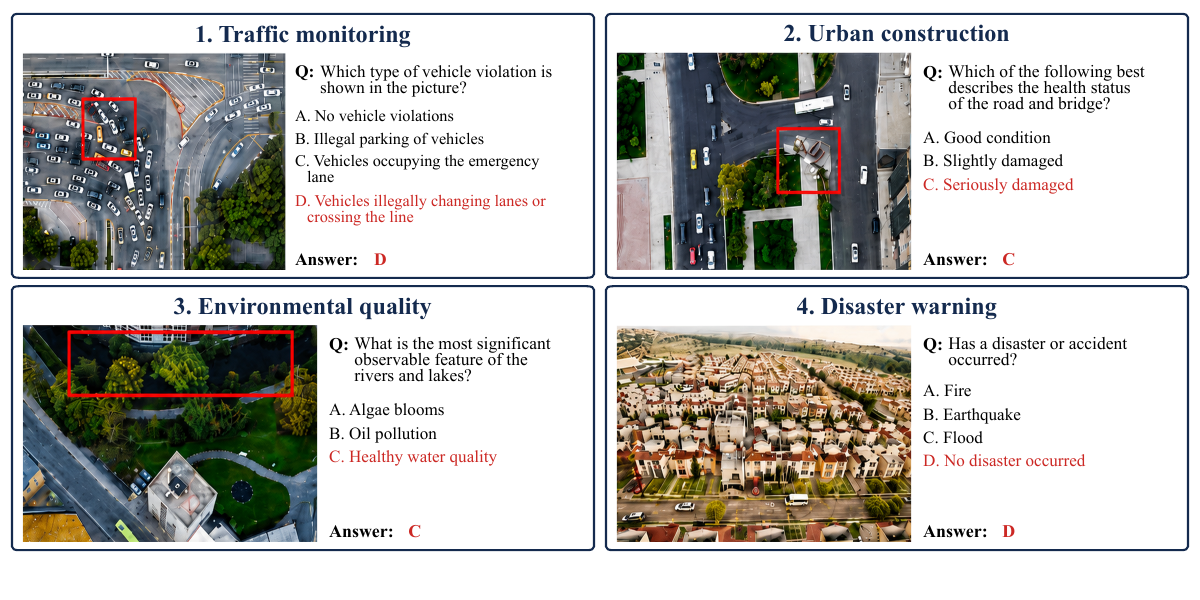} 
\vspace{-20pt}
\caption{Overview of UMBench. Red boxes are manually added for visualization only and are not provided to models during evaluation. Each example illustrates a compact multiple-choice question that requires fine-grained visual reasoning rather than scene-level priors.}
\label{fig:dataset_overview}
\end{figure*}

In this work, fine-grained visual reasoning denotes questions that cannot be solved from a global scene category alone. A correct answer must be supported by fine-grained evidence, including small-object inspection, attribute judgment, spatial-relation reasoning, or explicit verification against a plausible scene-level prior. This distinction is important for urban management because visually similar scenes may encode different operational states, such as lawful traffic versus lane crossing, intact pavement versus local cracking, or normal water surfaces versus inundation.

UMBench comprises 583 UAV images covering urban environments, complementing large-scale RS vision-language resources \cite{zhang2023rs5m, liu2024remoteclip}. The processed images are stored in PNG format, with the longer side resized to 800 pixels while preserving aspect ratio; GSD metadata was not available from the processed files. The image pool includes intersections, parking lots, residential blocks, construction zones, rivers, parks, green spaces, and open fields. Rather than treating high pixel resolution as the primary contribution, UMBench emphasizes task-relevant evidence such as vehicles, lane markings, barriers, pavement texture, water surfaces, vegetation patterns, and object relations. Blurred, overexposed, ambiguous, and near-duplicate images were removed during quality control.

As illustrated in Fig.~\ref{fig:dataset_overview}, the benchmark organizes 2,733 QA pairs into four application groups: {Traffic Monitoring} evaluates vehicle behavior, congestion, parking use, and lane relations; {Urban Construction} examines buildings, roads, and construction conditions; {Environmental Quality} covers water, vegetation, waste, and air-quality evidence; and {Disaster Warning} targets inundation, collapse, fire, obstruction, and infrastructure risks. A QA pair may receive multiple application tags, yielding 3,305 cumulative tag assignments (TM/UC/EQ/DW: 1,129/755/992/429).

Each instance is a multiple-choice question designed to resist answers based on scene-level priors. Annotators first identified a visually verifiable evidence region, formulated a question whose answer depends on that region, and then constructed scene-compatible distractors that differ in fine-grained visual evidence. All QA pairs were manually curated and reviewed by an expert team for visual verifiability, distractor plausibility, and ambiguity. Disagreements were resolved through consensus discussion, and ambiguous cases with unclear evidence or multiple plausible answers were revised or removed. Since labels were finalized through consensus-based expert curation, we report this quality-control protocol rather than an independent inter-annotator agreement score. The complete set is reserved for zero-shot evaluation without task-specific training.

\section{EVIDENCE-FIRST COT METHOD}

As discussed, current MLLMs may rely on scene-level priors when analyzing complex remote sensing imagery, allowing macroscopic context to override fine-grained visual evidence. Inspired by CoT and multimodal reasoning prompts \cite{wei2022chain, zhang2023multimodalcot}, we propose a lightweight, training-free evidence-first Chain-of-Thought prompting method. Instead of asking the model to reason freely, the method imposes an explicit evidence bottleneck: the model must first ground the question in visible local cues, then suppress scene-level priors, and finally map the verified evidence to a constrained option. This ordering is designed for UMBench, where distractors are often semantically plausible at the scene level but inconsistent with small visual details.

\begin{figure}[!h]
\centering
\fbox{
\parbox{0.9\linewidth}{
\small
\textbf{Input Configuration:} $\langle Image \rangle$, $\langle Question \rangle$, $\langle Choices \rangle$ \\
\textbf{Instruction Template:} \\
\texttt{[Visual Grounding]} \\
\textit{Inspect fine visual details first.} \\
\texttt{[Prior Suppression]} \\
\textit{Do not answer from scene type or semantic plausibility alone. Briefly cite concrete image evidence, then choose the best option.} \\
\texttt{[Output Formatting]} \\
\textit{End with: Final Answer: \textless Option\_Letter\textgreater}
}
}
\caption{Evidence-first prompting template. The model grounds local cues, suppresses unsupported scene-level priors, and returns a standardized option.}
\label{fig:prompt_framework}
\end{figure}

This design can be viewed as a compact \textit{verify-then-answer} framework, as summarized in Fig.~\ref{fig:prompt_framework}. It requires no fine-tuning or access to model weights, and can therefore be applied to both open-source and commercial models. The framework contains three modules. \textit{Visual Grounding} asks the model to inspect local objects, attributes, and spatial relations before using high-level semantics. \textit{Prior Suppression} weakens answers supported only by scene type or common co-occurrence patterns. \textit{Standardized Output} constrains the response to a final option letter, making evaluation robust to verbose outputs. 

The ordered stages prevent a fluent but unsupported conclusion from bypassing visual verification. Because the image, options, decoding configuration, and model parameters remain unchanged, the method isolates the effect of evidence-oriented prompting without introducing additional supervision or task-specific training.

\section{EXPERIMENTS AND RESULTS}

\subsection{Experimental Setup}
We evaluate MLLMs under two complementary access settings. In the \textit{local setting}, three compact open-source models are deployed under the same preprocessing and inference environment, providing a reproducible test of the method on resource-conscious systems. In the \textit{cloud setting}, three commercial models are queried without access to their weights, testing whether the same prompt-level intervention transfers to black-box services. All experiments are zero-shot and use identical images, questions, options, and decoding settings across models; only the prompting instruction changes. All models receive the processed PNG images without red-box annotations: local models use their official image processors and default visual-token settings, while commercial models receive the same images through provider APIs. We use fixed decoding settings when available and report exact top-1 accuracy on the full benchmark. Outputs without a valid final option are counted as incorrect.

We additionally compare direct answering with standard zero-shot CoT, which appends ``Let's think step by step'' to elicit intermediate reasoning \cite{kojima2022large}, and our evidence-first CoT. This control distinguishes gains attributable to generic, longer reasoning from those obtained by explicitly grounding the decision in local visual evidence. All three variants use the same samples and require a final option letter for automatic matching.

\begin{figure}[!t]
\centering
\includegraphics[width=\columnwidth]{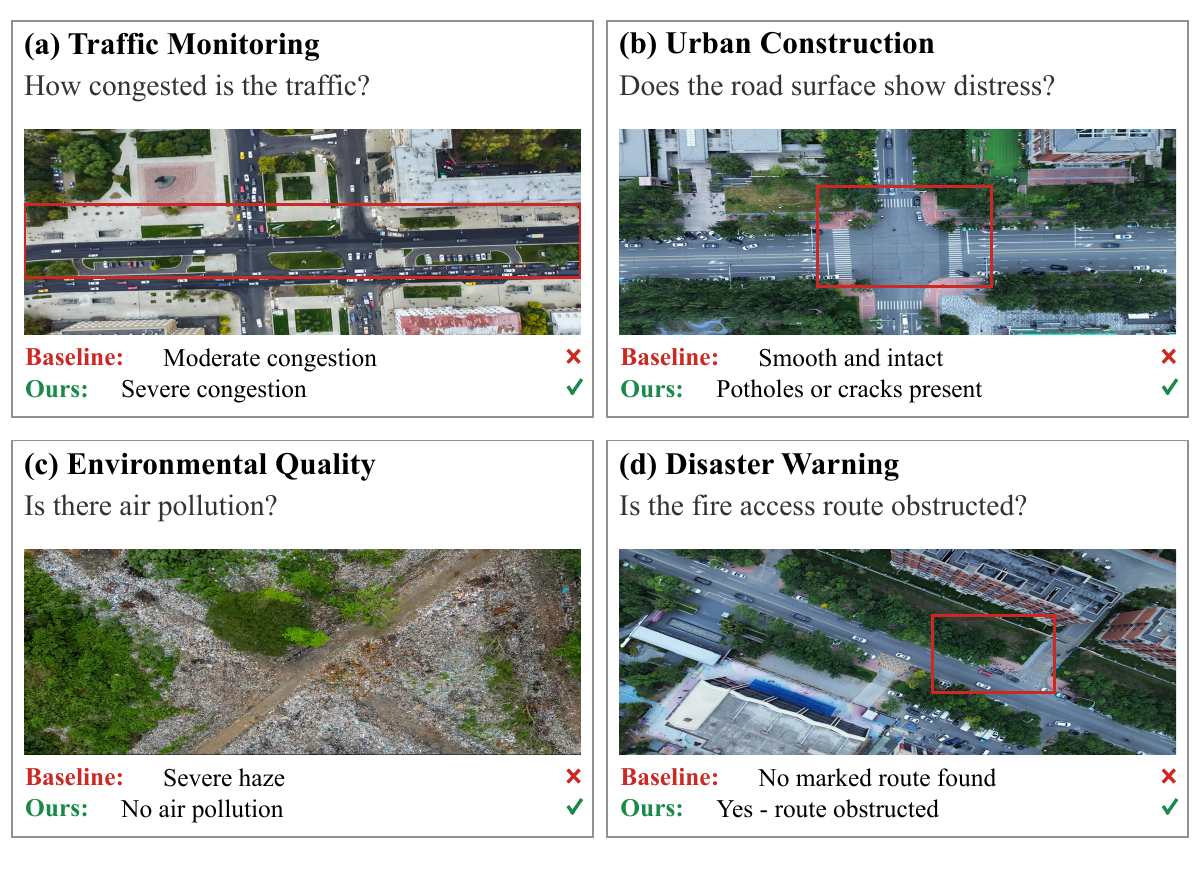}
\vspace{-20pt}
\caption{Representative corrections by evidence-first CoT on Qwen3-VL-8B. Direct answering tends to follow scene-level priors, while our prompt recovers the correct option from fine-grained visual evidence. Red boxes are for visualization only.}
\label{fig:qualitative_results}
\end{figure}

\subsection{Main Results and Analysis}

\renewcommand{\arraystretch}{1.00}
\begin{table}[!t]
\centering
\caption{Results of our method and open-source MLLMs on UMBench.}
\label{tab:local_models}
\resizebox{\columnwidth}{!}{
\begin{tabular}{l|ccccc}
\hline
\textbf{Method} & \textbf{TM} & \textbf{UC} & \textbf{EQ} & \textbf{DW} & \textbf{Avg} \\ \hline
Qwen3-VL-8B~\cite{bai2025qwen3vl} + Ours & \textbf{50.13} & 51.33 & 49.19 & \textbf{55.94} & \textbf{50.04} \\
Qwen3-VL-8B                       & 43.58 & \textbf{52.98} & 46.77 & 48.25 & 47.35 \\ \hline
InternVL3.5-8B~\cite{wang2025internvl35} + Ours & 45.88 & 46.75 & 47.48 & 54.55 & \textbf{46.62} \\
InternVL3.5-8B                   & 41.90 & 47.55 & 39.11 & 47.55 & 42.47 \\ \hline
Qwen2.5-VL-7B~\cite{bai2025qwen25vl} + Ours & 40.48 & 42.25 & 47.28 & 54.31 & \textbf{44.56} \\
Qwen2.5-VL-7B                    & 35.87 & 42.78 & \textbf{50.10} & 45.92 & 43.13 \\ \hline
\end{tabular}
}
\end{table}

\begin{table}[!t]
\centering
\caption{Results of our method and commercial MLLMs on UMBench.}
\label{tab:api_models}
\resizebox{\columnwidth}{!}{
\begin{tabular}{l|ccccc}
\hline
\textbf{Method} & \textbf{TM} & \textbf{UC} & \textbf{EQ} & \textbf{DW} & \textbf{Avg} \\ \hline
Gemini-3-Flash~\cite{googledeepmind2025gemini3flash} + Ours & \textbf{52.09} & \textbf{56.95} & \textbf{51.82} & 59.38 & \textbf{52.97} \\
Gemini-3-Flash & 51.90 & 55.63 & 50.10 & 57.11 & 51.58 \\ \hline
Grok-4.20~\cite{xai2026grok420} + Ours & 48.36 & 50.33 & 46.77 & \textbf{61.54} & 48.57 \\
Grok-4.20 & 46.77 & 54.17 & 33.67 & 42.19 & 44.97 \\ \hline
GPT-4o~\cite{openai2024gpt4o} + Ours & 49.07 & 53.25 & 47.78 & 59.21 & \textbf{49.27} \\
GPT-4o & 45.97 & 47.68 & 45.87 & 51.98 & 46.03 \\ \hline
\end{tabular}
}
\end{table}

Tables~\ref{tab:local_models} and~\ref{tab:api_models} summarize the results under two access settings: open-source MLLMs and commercial MLLMs. Table~\ref{tab:local_models} shows that evidence-first CoT improves the overall accuracy of all three open-source models, with the largest gain observed on InternVL3.5-8B. Table~\ref{tab:api_models} shows that the same training-free method can also be applied to commercial MLLMs, where Grok-4.20 obtains a clear overall improvement. 

Table~\ref{tab:cot_control} further isolates the effect of prompt structure on the three locally deployed open-source models. ZS-CoT raises InternVL3.5-8B from 42.47 to 44.31 and Qwen2.5-VL-7B from 43.13 to 43.57, whereas evidence-first CoT reaches 46.62 and 44.56, respectively. Thus, unconstrained step-by-step reasoning provides only part of the gain; explicitly verifying visual evidence is more effective overall. 


The direct-answer results show that current MLLMs still struggle with UMBench. Without explicit visual verification, models may select distractors that are consistent with scene-level priors while overlooking the fine-grained visual evidence required by the question. Evidence-first CoT reduces some of these errors by guiding models to inspect fine-grained visual evidence before selecting an option. The qualitative comparisons in Fig.~\ref{fig:qualitative_results} illustrate representative corrections across the four task groups, including congestion assessment, road distress, environmental evidence, and emergency-access obstruction.

Across these cases, evidence-first CoT does not add external knowledge, but redirects the model from scene-level priors to fine-grained visual evidence, such as vehicle density, pavement texture, pollution cues, and access-route continuity. This suggests that the method helps models use visual evidence more effectively, although prompting alone cannot fully solve fine-grained visual reasoning.

\begin{table}[!t]
\centering
\caption{Comparison of our method and ZS-CoT~\cite{kojima2022large} on open-source MLLMs using UMBench.}
\label{tab:cot_control}
\resizebox{\columnwidth}{!}{
\begin{tabular}{l|ccccc}
\hline
\textbf{Method} & \textbf{TM} & \textbf{UC} & \textbf{EQ} & \textbf{DW} & \textbf{Avg} \\ \hline
Qwen3-VL-8B + Ours & 50.13 & 51.33 & 49.19 & 55.94 & \textbf{50.04} \\
Qwen3-VL-8B + ZS-CoT~\cite{kojima2022large} & 45.26 & 47.68 & 45.97 & 55.48 & 46.51 \\ \hline
InternVL3.5-8B + Ours & 45.88 & 46.75 & 47.48 & 54.55 & \textbf{46.62} \\
InternVL3.5-8B + ZS-CoT~\cite{kojima2022large} & 42.60 & 46.62 & 46.27 & 50.35 & 44.31 \\ \hline
Qwen2.5-VL-7B + Ours & 40.48 & 42.25 & 47.28 & 54.31 & \textbf{44.56} \\
Qwen2.5-VL-7B + ZS-CoT~\cite{kojima2022large} & 39.42 & 41.85 & 45.26 & 54.78 & 43.57 \\ \hline
\end{tabular}
}
\end{table}

\section{CONCLUSION}

In this paper, we introduce UMBench, a remote sensing Urban Management Benchmark for evaluating fine-grained visual reasoning in MLLMs. UMBench contains 2,733 question-answer pairs across multiple urban management tasks and tests whether models can answer questions based on fine-grained visual evidence rather than scene-level priors. Our evaluations on locally deployed open-source models and cloud-hosted commercial models highlight two findings: 1) current MLLMs still struggle to consistently use fine-grained visual evidence; 2) the proposed evidence-first CoT prompting method improves performance without extra training or access to model weights, although the gains remain model- and task-dependent. Comparisons with ZS-CoT further show the value of explicit visual evidence inspection over generic step-by-step reasoning. These results suggest that future RS-specific MLLMs should strengthen fine-grained visual grounding, small-object perception, and attribute-level reasoning. UMBench provides a focused benchmark for studying fine-grained visual reasoning in urban remote sensing.

\section{AKNOWLEDGEMENT}

This research was supported by "the Key Research and Development Program of Heilongjiang Province under Grant 2026ZX01A07" and was funded by the National High-Level Young Talent Program Project.

\vspace{-0.6em}
\bibliographystyle{IEEEtranN}
\bibliography{references}

\end{document}